\def\year{2020}\relax
\documentclass[letterpaper]{article} 
\usepackage{aaai20}  
\usepackage{times}  
\usepackage{helvet} 
\usepackage{courier}  
\usepackage[hyphens]{url}  
\usepackage{graphicx} 
\usepackage{graphicx}  
\usepackage{times}
\usepackage{latexsym}
\usepackage{url}
\usepackage{epsfig}
\usepackage{graphicx}
\usepackage{amsmath}
\usepackage{amssymb}
\usepackage{multirow}
\usepackage{array}

\usepackage{mathtools}
\usepackage{bm}

\usepackage{xcolor, soul}
\sethlcolor{yellow}

\newcommand{\rbr}[1]{\left(#1\right)}

\newcommand{\hb}{\mathbf{h}}
\newcommand{\vb}{\mathbf{v}}
\newcommand{\eb}{\mathbf{e}}
\newcommand{\bb}{\mathbf{b}}
\newcommand{\Eb}{\mathbf{E}}
\newcommand{\Wb}{\mathbf{W}}
\newcommand{\ub}{\mathbf{u}}
\newcommand{\Dcal}{\mathcal{D}}

\newcommand*{\thetab}{\bm{\theta}}
\newcommand{\EE}{\mathbb{E}} 

\newcommand{\ngram}{\mathfrak{N}} 
\newcommand{\relevence}{\mathcal{R}} 
\newcommand{\coherence}{\mathcal{C}} 
\newcommand{\expressive}{\mathcal{E}} 
\newcommand{\model}{ReCo-RL } 
\newcommand{\mle}{MLE }
\newcommand{\bleu}{BLEU-RL }

\newcommand{\citet}[1]{\citeauthor{#1} \shortcite{#1}}

\title{What Makes A Good Story? Designing Composite Rewards for Visual Storytelling}

\author{Junjie Hu\textsuperscript{1}, Yu Cheng\textsuperscript{2}, Zhe Gan\textsuperscript{2}, Jingjing Liu\textsuperscript{2}, Jianfeng Gao\textsuperscript{3}, Graham Neubig\textsuperscript{1} \\
\textsuperscript{1}Carnegie Mellon University, \textsuperscript{2}Microsoft Dynamics 365 AI Research, \textsuperscript{3}Microsoft Research \\
\{junjieh, gneubig\}@cs.cmu.edu, \{yu.cheng, zhe.gan, jingjl, jfgao\}@microsoft.com}
 
\begin{document}

\maketitle

\begin{abstract}
Previous storytelling approaches mostly focused on optimizing traditional metrics such as BLEU, ROUGE and CIDEr. In this paper, we re-examine this problem from a different angle, by looking deep into what defines a natural and topically-coherent story. To this end, we propose three assessment criteria: \emph{relevance}, \emph{coherence} and \emph{expressiveness}, which we observe through empirical analysis could constitute a ``high-quality'' story to the human eye. We further propose a reinforcement learning framework, ReCo-RL\footnote{Codes are available in \url{https://github.com/JunjieHu/ReCo-RL}}, with reward functions designed to capture the essence of these quality criteria. Experiments on the Visual Storytelling Dataset (VIST) with both automatic and human evaluation demonstrate that our ReCo-RL model achieves better performance than state-of-the-art baselines on both traditional metrics and the proposed new criteria.
\end{abstract}

\section{Introduction}
There has been a recent surge of interest in enabling machines to understand the semantics of complex visual scenarios and depict visual objects/relations with natural language. One main line of research is grounding the visual concepts of a single image to textual descriptions, known as image captioning~\cite{fang2015captions,vinyals2015show,you2016image}. Visual storytelling~\cite{huang2016visual} takes one step further, aiming at understanding photo streams and generating a sequence of sentences to describe a coherent story. 

Most existing visual storytelling methods focus on maximizing data likelihood \cite{yu2017hierarchically}, topic consistency \cite{huang2018hierarchically},  or expected rewards by imitation learning \cite{xinwang-wenhuchen-ACL-2018}. However, maximizing data likelihood or implicit rewards does not necessarily optimize the quality of generated stories. In fact, we find that simply optimizing on standard automatic evaluation metrics may even hurt the performance of story generation according to other assessments that are more important to the human eye.  

\begin{figure*} 
\centering
\raisebox{0\height}{\includegraphics[width=0.945\textwidth]{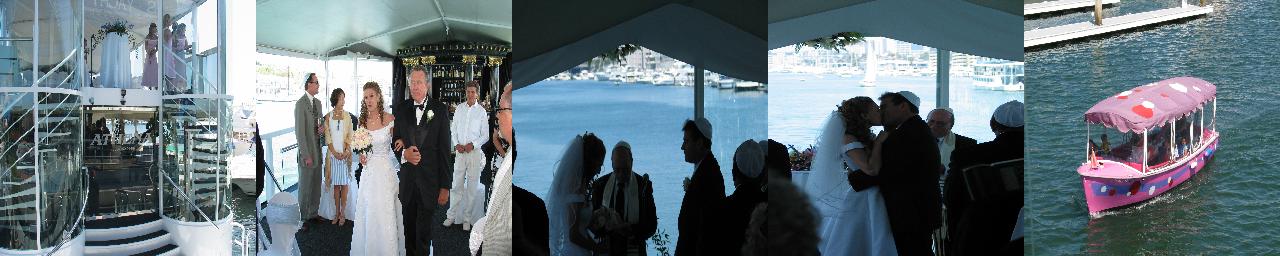}} \\
{\small
\begin{tabular}{|p{0.1\textwidth-2\tabcolsep}||p{0.835\textwidth-2\tabcolsep}|}  \hline
BLEU-RL & it was a great time at the \hl{wedding}. there was a lot of the \hl{wedding}. {\color{blue}it was a great time} at the reception. {\color{blue}it was a great time} to the end of the day. at the end of the \hl{boat}, we went to the \hl{boat}. \\ \hline 
Reference & a \hl{wedding} is getting ready to happen . there is \hl{flower girls} waiting the bride . \hl{father} is bring \hl{her} out . the \hl{bride} and \hl{groom} is getting married now . the \hl{bride} and \hl{groom} are kissing each other . this \hl{wedding} is taking place on a \hl{boat} . \\ \hline 
\end{tabular}}
\caption{Comparison between a story generated by the BLEU-RL model that is trained to optimize BLEU and human-written reference. Words in yellow indicate that there are more fine-grained concepts in the human-written reference than the model-generated one. The two segments in blue show an undesired repeating pattern in the output from the model.} \label{tab:example}
\end{figure*}

In this paper, we revisit the visual storytelling problem by asking ourselves the question: \textit{what makes a good story?} Given a photo stream, the first and foremost goal should be telling a story that accurately describes the objects and the concepts that appear in the photos. This can be termed as the \textit{``Relevance''} dimension. Secondly, the created story should read smoothly. In other words, the consecutive sentences should be semantically and logically coherent with each other, instead of being mutually-independent sentences describing each photo separately. This can be termed as the \textit{``Coherence''} dimension. Lastly, to tell a compelling story that can vividly describe the visual scenes and actions in the photos, the language used for creating the story should contain a rich vocabulary and diverse style. We call this the \textit{``Expressiveness''} dimension. 

Most existing storytelling approaches that optimize on BLEU or CIDEr do not perform very well on these dimensions. As shown in Figure \ref{tab:example}, compared with the model-generated story, the human-written one is more semantically relevant to the content of the photo stream (e.g., describing more fine-grained visual concepts such as ``flower girls''), more structurally coherent across sentences, and more diversified in language style (e.g., less repetition in pattern such as ``great time''). 

Motivated by this, we propose a reinforcement learning framework with composite reward functions designed to encourage the model to generate a relevant, expressive and coherent story given a photo stream. The proposed ReCo-RL (Relevance-Expressiveness-Coherence through Reinforcement Learning) framework consists of two layers: a high-level decoder (i.e., manager) and a low-level decoder (i.e., worker). The manager summarizes the visual information from each image into a goal vector, by taking into account the overall story flow, the visual context, and the sentences generated for previous images. Then it passes on the goal vector to each worker, which generates a word-by-word description for each image, guided by the manager's goal.

The proposed model consists of three quality evaluation components. The first \textit{relevance function} gives a high reward to a generated description that mentions fine-grained concepts in an image. The second \textit{coherence function} measures the fluency of a generated sentence given its preceding sentence, using a pre-trained language model. The third \textit{expressiveness function} penalizes phrasal overlap between a generated sentence and its preceding sentences. The framework aggregates these rewards and optimizes with the REINFORCE algorithm~\cite{williams1992simple}. Empirical results demonstrate that \model can achieve better performance than state-of-the-art baselines. Our main contributions can be summarized as follows:
\begin{itemize}
        \item We propose three new criteria to assess the quality of text generation for the visual storytelling task. 
        \item We propose a reinforcement learning framework, ReCo-RL, with composite rewards designed to align with the proposed criteria, using policy gradient for training.
    \item We provide quantitative analysis, qualitative analysis, and human evaluation to demonstrate the effectiveness of our proposed model. 
\end{itemize}

\begin{figure*}
    \centering
    \includegraphics[width=0.9\textwidth]{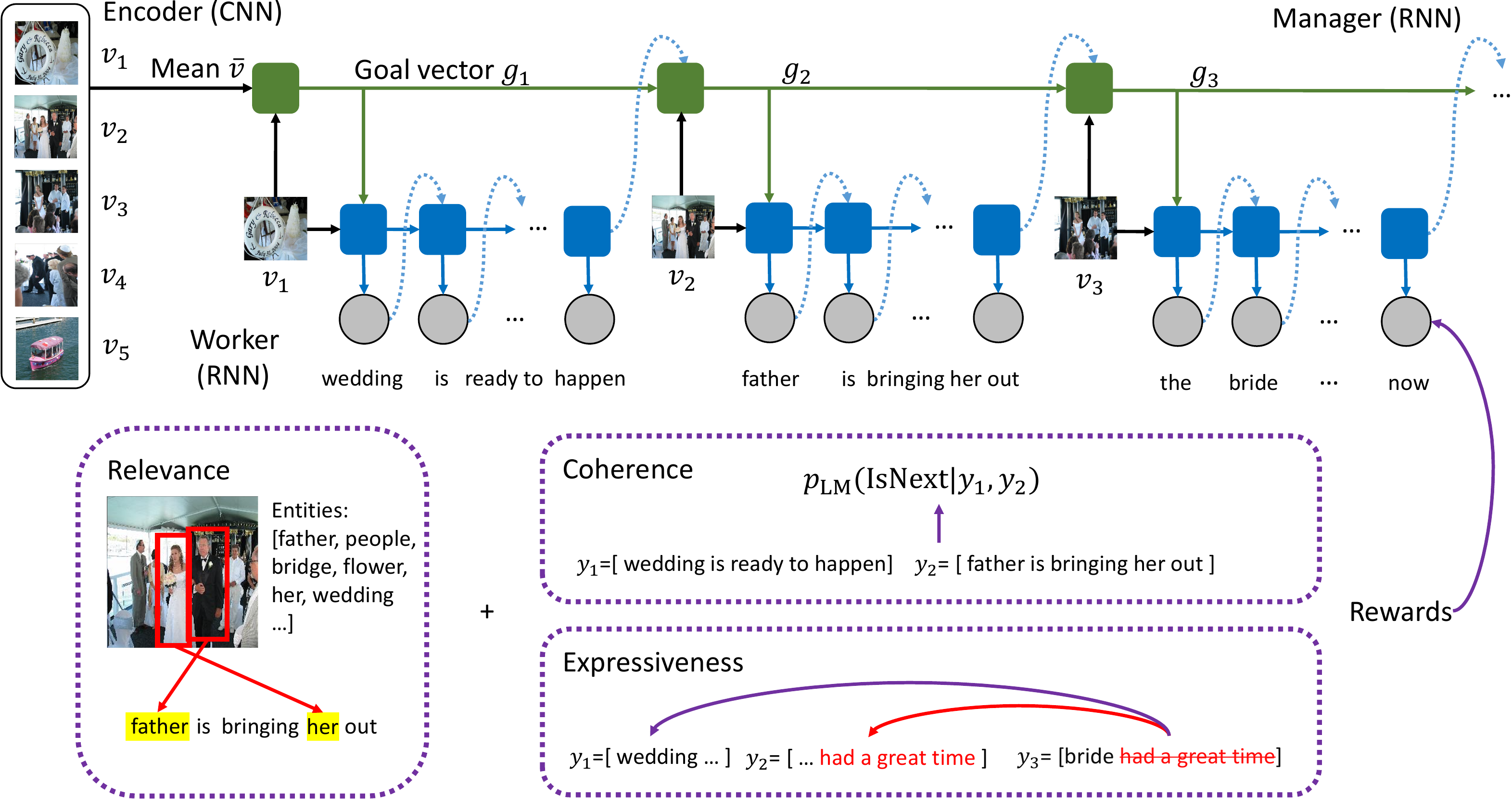}
    \caption{Model architecture and three rewards. Words highlighted in yellow show relevant concepts in the image. }
    \label{fig:my_label}
\end{figure*}

\section{Related Work}
\paragraph{Visual Storytelling} is a task where given a photo stream, the machine is trained to generate a coherent story in natural language to describe the photos. Compared with visual captioning tasks \cite{vinyals2015show,krishna2017dense,RennieMMRG17,gan2017semantic}, visual storytelling requires capabilities in understanding more complex visual scenarios and generating more structured expressions. Pioneering work has used sequence-to-sequence model on this task~\cite{park2015expressing}. \citet{huang2016visual} provided the benchmark dataset VIST for this task. \citet{yu2017hierarchically} have shown promising results on VIST with a multi-task learning algorithm for both album summarization and sentence generation. 

Recent efforts have explored REINFORCE training, by learning an implicit reward function~\cite{xinwang-wenhuchen-ACL-2018} to mimic human behavior or injecting a topic consistency constraint during training~\cite{huang2018hierarchically}. \citet{DBLP:conf/aaai/WangFTLM18} proposed a hierarchical generative model to create relevant and expressive narrative paragraphs. To improve the structure and diversity, \citet{NIPS2018_7426} reconciled a traditional retrieval-based method with a modern learning-based method to form a hybrid agent. Notably, these studies did not directly (or explicitly) examine what accounts for a good story to the human eye, which is the main focus of our work. 

\paragraph{Text Generation} State-of-the-art text generation methods use encoder-decoder architectures for sequence-to-sequence learning \cite{Bordes2016LearningEG,Sutskever2014SSL}. To better model structured information, hierarchical models have been proposed \cite{P15-1107}. Follow-up work tried to overcome exposure bias resulting from MLE training \cite{Bengio:2015:SSS:2969239.2969370,NIPS2016_6099}. In recent years, reinforcement learning (RL) has gained popularity in many tasks \cite{Ranzato2015SequenceLT}, such as image captioning \cite{RennieMMRG17}, text summarization \cite{DBLP:journals/corr/PaulusXS17} and story plot generation \cite{Tambwekar19}. Other techniques such as adversarial learning \cite{seqgan,dai17,zhang2017adversarial}, inverse reinforcement learning (IRL) \cite{NIPS2016_6391} and pre-training~\cite{chen2019uniter} have also been applied. Compared with previous work, we define explicit rewards for the visual storytelling task and propose a reinforcement learning framework to optimize them.

Meanwhile, how to assess the quality of generated text still remains a major challenge. BLEU~\cite{papineni2002bleu} and METEOR~\cite{banerjee2005meteor} are widely used in machine translation. CIDEr~\cite{vedantam2015cider} and SPICE~\cite{anderson2016spice} are used for image captioning. ROUGE-L~\cite{lin2004rouge} is used for evaluating text summarization. However, these metrics all have limitations in evaluating natural language output, as there exists a large gap between automatic metrics and assessment by humans. There have been some recent studies on more natural assessment for text generation tasks, such as evaluating on structuredness, diversity and readability \cite{plainandwrite,dai17,chen2018fast,xinwang-wenhuchen-ACL-2018}, although these studies do not explicitly consider relevance between a stream of images and a story for the task of visual storytelling. Similar to these studies, we argue that the aforementioned automatic metrics are not sufficient to evaluate the visual storytelling task, which requires high readability and naturalness in generated stories.

\section{Approach}

\subsection{Notation}
Given a stream of $n$ images, we denote their features extracted by a pre-trained convolutional neural network as a sequence of vectors $V\equiv[\vb_1,\cdots, \vb_n]$. The reference descriptions are denoted as a sequence of sentences $Y\equiv[y^*_1,\cdots, y^*_n]$, where $y^*_i$ is a sequence of word indices that depicts the $i$-th image. We define a dataset of input-output pairs as $\Dcal=\{(V, Y)\}$. Based on the reference $i$-th image, our model generates the corresponding sentence $y_i$, where $y_i^t$ denotes the $t$-th word in $y_i$. We denote $\Eb$ as the word embedding matrix, and $\eb_i^t=\Eb[y_i^t]$ as the word embedding of $y_i^t$. We denote the hidden state of the manager and the worker as $\hb_{M,i}$ and $\hb_{W,i}^t$, respectively. We use a bold letter to denote a vector or matrix, and use a non-bold letter to denote a sequence or a set.

\subsection{Model Architecture}
\label{sec:model}
\paragraph{The Encoder} module consists of a pre-trained convolutional neural network which extracts deep visual features from each image, with ResNet-101~\cite{he2016deep}. The encoder obtains the overall summary of a photo stream by averaging the visual features of all the images, i.e., $\bar{\vb}=\frac{1}{n}\sum_i \vb_i$.

\paragraph{The Manager} module in our model is a Long Short-term Memory (LSTM) network, which captures the consistency of the generated story at the sentence-level. When depicting one image of a photo stream, the manager should take into account three aspects: 1) the overall flow of the photo stream; 2) the context information in the current image; and 3) the sentences generated from previous images in the photo stream. To do so, for each image in the $i$-th step, the manager takes as input the features of the whole image sequence $\bar{\vb}$, the features of the $i$-th image $\vb_i$, and the worker's last hidden state $\hb_{W,i-1}^T$ from the previous image. The manager then predicts a hidden state as the goal vector. 
\begin{align}
    \hb_{M,i} = \text{LSTM}_M\rbr{[\bar{\vb}; \vb_i; \hb_{W,i-1}^T], \hb_{M,i-1}} 
\end{align}
where $[;]$ denotes vector concatenation. The goal vector is then passed on to the worker, and the worker is responsible for completing the generation of word description based on the goal from the manager.

\paragraph{The Worker} module is a fine-grained LSTM network, which predicts one word at a time and controls the fluency of one sentence. Intuitively, the worker is guided by the goal from the manager, and focuses more on fine-grained context information in the current image. More specifically, when predicting one word at the $t$-th step, the worker takes as input the features of the $i$-th image $\vb_i$, the manager's goal vector $\hb_{M,i}$, and the word embedding of the previously generated word $\eb_i^{t-1}$. The worker then predicts a hidden state $\hb_{W,i}^t$ and applies a linear layer $f$ to approximate the probability of choosing the next word in Eq. (\ref{eq:prob}). 

\begin{align}
    &\hb_{W,i}^t = \text{LSTM}_W\rbr{[\vb_i; \hb_{M,i}; \eb_i^{t-1}], \hb_{W,i}^{t-1}} \\ \label{eq:prob}
    &p_{\thetab}(y_i^t| y_i^{1:t-1}, \vb_i, \bar{\vb}) = \text{softmax}(f(\hb_{W,i}^t)) \\
    &p_{\thetab}(y_i|\vb_i, \bar{\vb}) = \prod_t p_{\thetab}(y_i^t| y_i^{1:t-1}, \vb_i, \bar{\vb})
\end{align}

\subsection{Composite Rewards Design}
\label{sec:reward}
\paragraph{Relevance}
One way to measure the relevance between an image and its generated description is to ground the entities mentioned in the description to corresponding bounding boxes in the image. However, a straightforward way of comparing the n-gram overlap between the reference sentence and the generated sentence (e.g., BLEU or METEOR) treats each word in the sentence equally, without taking into account the semantic relevance of the words to the image. 

To tackle this limitation, we propose to measure the semantic similarity between entities mentioned in the reference and generated sentences. More specifically, we are given a set of $K$ reference sentences $Y_i^* = \{y_{i,k}^*\}_{k=1}^K$ for the $i$-th image. We then extract a set of entities $O^{Y_i^*}$ mentioned in its reference sentences $Y_i^*$ with a Part-Of-Speech (POS) tagger, and count the frequency of the entities in its reference sentences as $C(o, Y_i^*), \forall o\in O^{Y_i^*}$. The normalized frequency of an entity is computed by dividing by the sum of the frequency of all entities of in $O^{Y_i^*}$ in Eq.~(\ref{eq:norm_freq}). 
\begin{align} \label{eq:norm_freq}
    F(o, Y^*_i) &= \frac{C(o, Y^*_i)}{\sum_{o'\in O^{Y^*_i}} C(o', Y^*_i)} 
\end{align}
Similarly, we extract all the entities mentioned in an n-gram of a hypothesis $y_i$ sampled by the model, and denote the hypothesis n-gram as $\ngram$ and its entity set as $O^\ngram$. To measure the relevance of each hypothesis n-gram with respect to the key concepts in an image, we compute the relevance weight of an n-gram in Eq.~(\ref{eq:ngram_weight}). 
\begin{align}  \label{eq:ngram_weight}
    W^\ngram &= 1 + \beta\sum_{o \in O^\ngram \cap O^{Y^*_i}} F(o, Y^*_i) 
\end{align}
If a hypothesis n-gram contains any key entities in $O^{Y_i^*}$, $W^\ngram$ is greater than 1, which distinguishes it from other n-grams that do not ground to any bounding objects in the image. Notice that the weight is proportional to the number of key entities in $O^{Y_i^*}$ and the entity frequency in the reference sentences $Y_i^*$. Intuitively, the more entities an n-gram contains, the more bounding objects in the image this n-gram grounds to. If an entity is mentioned by multiple annotators in the reference sentences, the weight of mentioning this entity in the hypothesis should be high.

Inspired by the modified n-gram precision in the BLEU score calculation, we aim to avoid rewarding multiple identical n-grams in the hypothesis. To this end, we count the maximum number of times an n-gram exists in any single reference sentence in Eq.~(\ref{eq:c_max}), and clip the count of each hypothesis n-gram by its maximum reference count in Eq.~(\ref{eq:c_clip}). We then compute the weighted precision of all the n-grams in the hypothesis $y_i$ in Eq.~(\ref{eq:weighted_prec}).
\begin{align} \label{eq:c_max}
    & C_\text{max}(\ngram, Y^*_i) = \max_{y^*_{i,k}\in Y^*_i} C(\ngram, y^*_{i,k}) \\ \label{eq:c_clip}
    & C_\text{clip}(\ngram, y_i)  = \min\{C(\ngram, y_i), C_\text{max}(\ngram, Y_i)\} \\ \label{eq:weighted_prec}
    &P_n = \frac{\sum_{\ngram \in y_i} C_\text{clip}(\ngram, y_i) \cdot W^\ngram }{ \sum_{\ngram' \in y_i} C(\ngram', y_i) \cdot W^{\ngram'} }
\end{align}
 The relevance score of a sampled hypothesis with respect to the key concepts of an image is computed as the product of a brevity penalty and the geometric mean of the weighted n-gram precision in Eq.~(\ref{eq:relevence}). In our implementation, we consider unigram and bigram, i.e., $n=2$, since most entities only contain one or two words.
\begin{align} \label{eq:relevence} 
    \relevence(y_i) &= \text{BP} \rbr{\prod_{i=0}^n P_n}^{\frac{1}{n}} \\
    \text{BP}& =\exp \rbr{\min\rbr{1- \frac{r}{|y_i|}, 0}} 
\end{align}

\paragraph{Coherence}
A coherent story should organize its sentences in a correct sequential order and preserve the same topic among adjacent sentences. One way to measure coherence between two sentences is a sentence coherence discriminator that models the probability of two sentences $y_{i-1}$ and $y_i$ being continuous in a correct sequential order as well as containing the same topic. 

To this end, we leverage a language model with a next-sentence-prediction objective, as was explored in~\citet{devlin2018bert}. We first construct a sequence by concatenating two sentences $y_{i-1}$ and $y_i$ decoded by our model, and get the sequence representation using a pre-trained language model. Then, we apply a linear layer to the sequence representation followed by a $\tanh$ function and a softmax function to predict a binary label, which indicates whether the second sentence is the sentence that follows the first one.
\begin{align}
    \ub_{i-1,i} & = \text{LM}(y_{i-1}, y_{i}) \\
    p_\text{LM}(s|y_{i-1}, y_{i}) & = \text{softmax} \rbr{\tanh \rbr{\Wb \ub_{i-1,i} + \bb}} \\
    \coherence(y_i) & = p_\text{LM}(s=0|y_{i-1}, y_{i})
\end{align}
where $s=0$ indicates $y_{i}$ is the sentence that follows $y_{i-1}$. 
\paragraph{Expressiveness}
An expressive story should contain diverse phrases to depict the rich content of a photo stream, rather than repeatedly using the same words. To capture this expressiveness, we keep track of already-generated n-grams, and punish the model when it generates repeated n-grams. 

To this end, we propose a diversity reward which measures the n-gram overlap between the current sentence $y_{i}$ and previously decoded sentences $\{y_1,\cdots, y_{i-1}\}$. More specifically, we first regard all the preceding sentences $\{y_1,\cdots, y_{i-1}\}$ as the reference sentences to the current sentence $y_{i}$, and compute the BLEU score of the current sentence compared to the reference sentences. Finally we substract this value from 1 as the expressiveness reward in Eq.~(\ref{eq:expressiveness}). Intuitively, if the current sentence contains more identical n-grams as any one of preceding decoding sentences, the BLEU score of the current sentence with respect to that already-generated sentence would be high, thus the story is lack of expressiveness when adding the current decoding sentence. In our implementation of BLEU in Eq.~(\ref{eq:expressiveness}), we only consider the precision of bigram, trigram and 4-gram, since we want to focus on repeated phrases that have more than one word. 
\begin{align} \label{eq:expressiveness}
    \expressive(y_i) = 1 - \text{BLEU}(y_{i},\{y_1,\cdots, y_{i-1}\}) 
\end{align}

\subsection{Training}
\label{sec:train}
We first train our proposed model using maximum likelihood estimation (MLE), and then continue training the model using REINFORCE algorithm together with an MLE objective.
\paragraph{Maximum Likelihood Estimation} seeks an optimal solution $\thetab^*$ by minimizing the negative log-likelihood of predicting the next word over batches of training observations in Eq.~(\ref{eq:mle}). We apply stochastic gradient descent to update the model parameters on each mini-batch of data $\Dcal'$ in Eq.~(\ref{eq:sgd_mle}).
\begin{align} \label{eq:mle}
    J_\text{MLE}(\thetab, \Dcal') &= \!\!\!\! \sum_{Y,V \in \Dcal'} \! \sum_{i=1}^n \! -\log p_{\thetab} (y_i^*|\vb_i,\bar{\vb}) \\ \label{eq:sgd_mle}
    \thetab & \leftarrow \thetab + \eta \frac{\partial J_\text{MLE}(\thetab, \Dcal') }{ \partial \thetab}
\end{align}
where $\eta$ is the learning rate.

\paragraph{REINFORCE} \cite{williams1992simple} is able to learns a policy by maximizing an arbitrary expected reward in Eq.~(\ref{eq:rl}). This makes it possible to design reward functions specifically for the visual storytelling task. We compute the weighted sum of the aforementioned three reward functions, to encourage the model to focus on those key aspects of a good story and control the generation quality of the sentences.
\begin{align} \label{eq:rl}
     J_\text{RL}(\thetab) &= \sum_{Y,V\in\Dcal'}\!\! \EE_{y_i\sim \pi_i}\left[(b - r(y_i)) \log \pi_i\right]   \\
     r(y_i) & = \lambda_R \relevence(y_i) + \lambda_C \coherence(y_i) + \lambda_E \expressive(y_i)  
\end{align}
where $\pi_i\equiv p_{\thetab}(y_i|\vb_i,\bar{\vb})$ is the policy, and $b$ is a baseline that reduces the variance of the expected rewards, $\lambda_R$, $\lambda_C$ and $\lambda_E$ are the weights of the three designed rewards. In our implementation, we sample $H$ hypotheses generated by the current policy $\pi_i$ for the $i$-th image, and approximate the expected rewards with respect to the empirical distribution $\pi_i$. We compute the baseline by using the average reward of all the sampled hypotheses, i.e., $b=\frac{1}{H}\sum_{y_i\sim\pi_i} r(y_i)$. 

Rather than starting from a random policy model, we start from a model pre-trained by the MLE objective, and continue training the model jointly with MLE and REINFORCE objectives on each mini-batch $\Dcal'$ in Eq.~(\ref{eq:mle_rl}), following~\citet{Ranzato2015SequenceLT}.
\begin{align}\label{eq:mle_rl}
    \!\!\!\!\!\thetab  \! \leftarrow \! \thetab + \eta_1 \frac{\partial J_\text{MLE}(\thetab, \Dcal') }{\partial \thetab} +\eta_2 \frac{\partial J_\text{RL}(\thetab, \Dcal') }{ \partial \thetab}
\end{align}

\begin{table}[t!]
\resizebox{0.46\textwidth}{!}{
\centering
\begin{tabular}{c|ccccc}
\hline
Method & METEOR & ROUGE & CIDEr & BLEU-4 & SPICE\\ \hline \hline 
AREL   & \textbf{35.2}   & 29.3    & \textbf{9.1}     & 13.6  & \textbf{8.9}\\
HSRL & 30.1 & 25.1 & 5.9 & 9.8  & 7.5\\\hline 
\mle             & 34.8   & 30.0    & 7.2     & 14.3  & 8.5\\ 
\bleu  &  \textbf{35.2}     &  \textbf{30.1}   & 6.7     & \textbf{14.4}  & 8.3\\ \hline
\model   &  33.9     &  29.9   & 8.6     &  12.4  & 8.3\\\hline 
\end{tabular}}
\caption{Comparison between different models on METEOR, ROUGE-L, CIDEr, BLEU-4 and SPICE.}
\label{tab:auto_eval}
\end{table}

\begin{table*}
\resizebox{1\textwidth}{!}{
\begin{tabular}{@{}c@{}|@{~}c@{~}c@{}c@{~}c@{~}|@{~}c@{~}c@{}c@{~}c@{~}|@{~}c@{}c@{}c@{~}c@{~}|@{~}c@{~}c@{}c@{~}c@{~}}
\hline
Aspects & AREL & \model & Tie & Agree & HSRL & \model & Tie & Agree & \mle & \model & Tie & Agree & \bleu & \model & Tie & Agree \\
\hline \hline
R & 27.6\% & 62.2\% & 10.2\% & 0.72 & 36.1\% & 53.8\% & 10.1\% & 0.74 & 27.0\% & 64.1\% & 8.9\%& 0.49 & 17.6\% & 74.5\% & 7.9\%& 0.78 \\
C & 31.3\% & 58.7\% & 10.0\% & 0.78 & 38.0\% & 51.9\% & 10.1\% & 0.80 & 34.3\% & 57.7\% & 8.0\%& 0.53 & 18.9\% & 72.3\% & 8.8\%& 0.71 \\
E & 32.4\% & 58.6\% & 9.0\% & 0.68 & 38.6\% & 53.3\% & 8.1\% & 0.72 & 30.5\% & 61.0\% & 8.5\% & 0.55 & 19.5\% & 71.5\% & 9.0\% & 0.62 \\
\hline\hline
\end{tabular}}
\caption{Pairwise human comparison between \model and three methods on three quality aspects (\textbf{R}: Relevance, \textbf{C}: Coherence, \textbf{E}: Expressiveness). For each pairwise comparison, the first three columns indicate the percentage that turkers prefer one system outputs over the other one, and turkers think both stories are of equal quality. The last column is the Fleiss' kappa~\cite{fleiss1971mns} which is a statistical measure of inter-rater consistency. Agreement scores in the range of $[0.6,0.8]$ show substantial agreement between multiple turkers. }\label{tab:human_eval}
\end{table*}

\section{Experiment}
\subsection{Dataset and Baseline} 
\textbf{Dataset}: The VIST dataset \cite{huang2016visual} used in our evaluation consists of 10,117 Flickr albums with 210,819 unique photos. Each sample contains one story that describes 5 selected images from a photo stream, and the same album is paired with 5 different stories as references. The split is similar to previous work, with 40,098 samples for training, 4,988 for validation and 5,050 for testing. The vocabulary size of VIST is 12,977. The released data was processed by a name entity recognition (NER) tagger to solve the sparsity issue of low-frequence words. The name of a person, a location and an organization are replaced by [male]/[female], [location], and [organization], respectively. 

\noindent \textbf{Implementation Details}: The visual features are extracted from the last fully-connected layer of ResNet152 pretrained on ImageNet \cite{he2016deep}. The word embeddings of size 300 are uniformly initialized within $[-0.1,0.1]$. We use a 512-hidden-unit LSTM layer for both the manager and the worker modules. We apply dropout to the embedding layer and every LSTM layer with the rate of 0.3.  We set the hyper-parameters $\lambda_R = \lambda_C = \lambda_E = 1$ to assign equal weights to all the three aspects of the reward functions, and set $\eta_1 = \eta_2 = 1$ to balance both MLE and REINFORCE objectives during training. We use BERT~\cite{devlin2018bert} as our next sentence predictor and fine-tune the predictor on sentence pairs in the correct and random order in the VIST dataset. For negative sentence pairs, we randomly concatenate two sentences in two different albums to make sure that the topics of these sentences are different. 

\noindent \textbf{Baseline}: We compare our method with the following baselines:
(1) \textit{AREL} \cite{xinwang-wenhuchen-ACL-2018}\footnote{\url{https://github.com/eric-xw/AREL.git}}, an approach to learn an implicit reward with imitation learning;
(2) \textit{HSRL} \cite{huang2018hierarchically}\footnote{Codes are provided by the authors.}, a hierarchical RL approach that injects a topic consistency constraint during training. 
These two approaches achieved state-of-the-art results on VIST, and we follow the same parameter settings in the original papers. 

In addition, we also compare three variants of our model: (1) \textit{\mle} that uses MLE training in Eq.~(\ref{eq:sgd_mle}); (2) \textit{\bleu}~that is jointly trained by MLE and REINFORCE, using sentence-level BLEU as a reward; and (3) \textit{\model} that is jointly trained by MLE and REINFORCE, using the designed rewards in Eq.~(\ref{eq:mle_rl}). The decoding outputs generated are evaluated by the same scripts as \citet{xinwang-wenhuchen-ACL-2018}. 

\subsection{Quantitative Evaluation} 
Automatic metrics, including METEOR, CIDEr, BLEU-4, ROUGE-L and SPICE are used for quantitative evaluation. Table \ref{tab:auto_eval} summarizes the results of all the methods in comparison. Our models (MLE, \bleu and ReCo-RL) achieve competitive or better performance over the baselines on most metrics except CIDEr. Specially, \bleu achieves better performance in METEOR, ROUGE-L and BLEU-4, while \model improves the CIDEr score.

In addition to the automatic metrics, we can also use the designed reward functions to score each story generated by different methods. To evaluate the overall performance of one method at the corpus level, we average the reward scores of all stories generated by the method on the test set. Similar to the automatic evaluation metrics, we multiply the average reward scores by 100 and report the scaled results of all the methods on the test set in Table~\ref{tab:rewards}. Our proposed \model method outperforms all the start-of-the-art methods and our variants (BLEU-RL, MLE) on all three quality aspects.

\begin{table}[!t]
\centering
\begin{tabular}{c|ccc}
\hline
Method  & Relevance      & Coherence      & Expressiveness \\ \hline\hline
HSRL    & 1.95           & 7.21           & 33.27          \\ 
AREL    & 3.27           & 9.90           & 34.98          \\ \hline
MLE     & 5.46           & 7.92           & 30.76          \\ 
BLEU-RL & 2.17           & 12.40          & 30.41          \\ \hline
ReCo-RL & \textbf{10.39} & \textbf{12.74} & \textbf{39.37} \\ \hline
\end{tabular}%
\caption{Comparison between different models on three rewards, i.e., Relevance, Coherence and Expressiveness.}
\label{tab:rewards}
\end{table}

\begin{figure*}[t!] 
\centering
{\small
\begin{tabular}{|p{0.077\textwidth}||p{0.41\textwidth}||p{0.41\textwidth}|}  \hline
Methods & \includegraphics[width=0.41\textwidth]{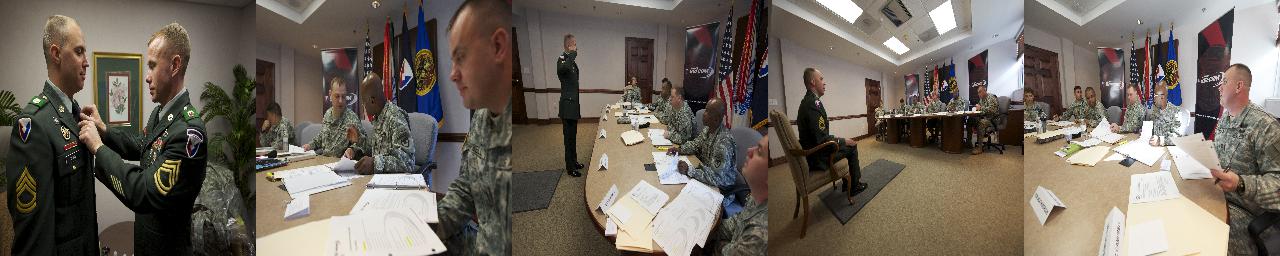} & \includegraphics[width=0.41\textwidth]{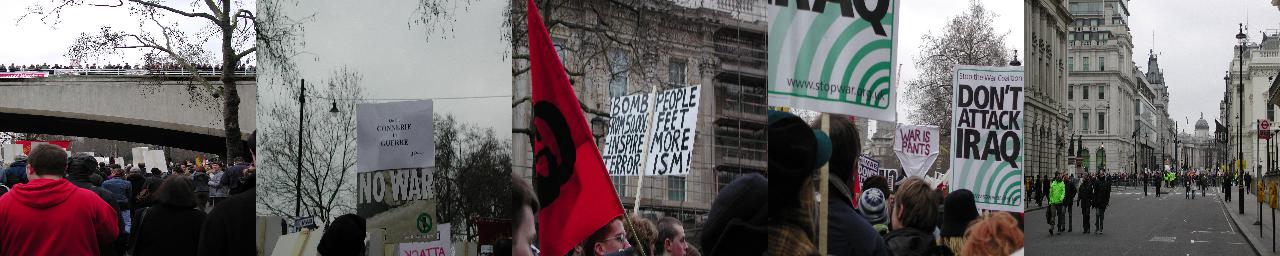} \\\hline
AREL & the officers of the \hl{military officers} are in charge of the \hl{military}. he was very proud of his speech. the \hl{meeting} was a great success. \underbar{the president of the company gave a} \underbar{speech to the \hl{audience}. we had a great time.} &  {\color{blue}there were a lot of people there. there was a lot of people there. there were many people there. there were a lot of people there.} the \hl{streets} were filled with \hl{people}. \\\hline 
HSRL & \underbar{i was so excited to see my \hl{new} \hl{team}. he was very happy} \underbar{to see the new professor.} i had a lot of time to talk about. {\color{blue}i had a great time. i had a great time.}  & we went to the local \hl{park} to celebrate the day. {\color{blue}there were many people dressed up in costumes. there were many \hl{people} dressed up in costumes. there were many \hl{people} there.} they had a great time and the \hl{crowd} was very excited. \\\hline 
\bleu & at the end of the day, the men were very proud of the \hl{military}. they had a lot of \hl{people} there. this is a picture of the \hl{meeting}. a group of \hl{people} had a great time. after the end of the day , we all had a lot of questions. & the \hl{crowd} gathered for the \hl{event}. there were many \hl{people} there . we had {\color{blue}a lot of \hl{people}} there . {\color{blue}a lot of \hl{people}} were protesting . the \hl{streets} were very tall . \\\hline
\model  & today was a picture of the \hl{military officer}, \hl{he} was ready to go to the organization. \hl{they} were very happy to see the awards ceremony. the \hl{speaker} was very excited to be able to talk about the \hl{meeting}. \hl{everyone} was having a great time to get together for the event after the ceremony. we all had a lot of \hl{people} there. &  the \hl{crowd} was very excited to see the \hl{protest}. we had a lot of \hl{people} there to get a \hl{sign} in the city. they were very proud of the \hl{flags}. i 'm glad to go to the \hl{rally}. many \hl{people} gathered around the \hl{street} to show the \hl{signs}. after the day, we went to location. \\ 
\hline 
\end{tabular} } 
\caption{Example stories generated by our model and the baselines. Words in yellow indicate entities appearing in the image, and words in blue show repetitive patterns. Pairs of sentences that describe different topics are annotated by an underline.} \label{fig:example}
\end{figure*}

\subsection{Human Evaluation}
Due to the subjective nature of the storytelling task, we further conduct human evaluation to explicitly examine the quality of the stories generated by all the models, through crowdsourcing using Amazon Mechanical Turk (AMT). 
Specifically, we randomly sampled 500 stories generated by all the models for the same photo streams. Given one photo stream and the stories generated by two models, three turkers were asked to perform a pairwise comparison and select the better one from the two stories based on three criteria: \textit{relevance}, \textit{coherence} and \textit{expressiveness}. 
The user interface of the evaluation tool also provides a neutral option, which can be selected if the turker thinks both outputs are equally good on one particular criterion. The order of the outputs for each assignment is randomly shuffled for fair comparison. Notice that in the pairwise human evaluation, each pair of system outputs for one photo stream was judged by a different group of three people. The total number of turkers for all photo streams is 862.

Table~\ref{tab:human_eval} reports the pairwise comparison between \model and three other methods. 
Based on human judgment, the quality of the stories generated by \model are significantly better than the \bleu variant on all dimensions, even though \bleu is fine-tuned to obtain comparative scores on existing automatic metrics.  Comparing with two strong baselines, AREL and HSRL, \model can still achieve better performance.  For each pairwise comparison between two model outputs, we also scored each story based on the number of votes from three turkers, and performed the Student's paired t-test between the scores of two systems. Our \model is significantly better than all baseline methods with $\rho<0.05$.

\begin{figure*}[t!]
\resizebox{0.98\textwidth}{!}{ 
\begin{tabular}{|p{0.1\textwidth}||p{0.58\textwidth}||c|c|c|c|} \hline
\multirow{6}{*}{Method} & \multirow{6}{*}{\includegraphics[width=0.58\textwidth]{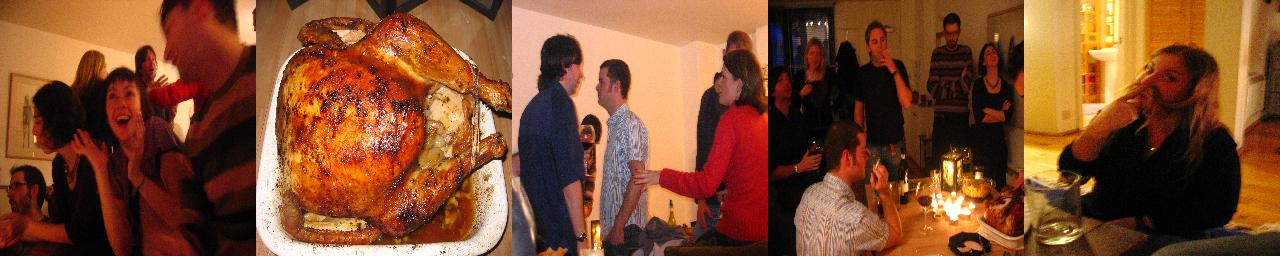}} & \multicolumn{4}{c|}{} \\
                        &                        &  \multicolumn{4}{c|}{Quality Metrics}          \\ 
                        &                        &  \multicolumn{4}{c|}{} \\  \cline{3-6} 
                        &                        &        &        &       &            \\ 
                        &                        &  R & C   & E     &   B         \\ 
                        &                        &        &        &       &            \\ \hline
\multirow{3}{*}{\bleu} & a group of \hl{friends} gathered together for \hl{dinner}. the \hl{turkey} was delicious. the guests were having {\color{blue}a great time. at the end of the night, we had a great time. at the end of the night, we had a great time.} & \multirow{3}{*}{2.47} & \multirow{3}{*}{11.06} & \multirow{3}{*}{37.10} & \multirow{3}{*}{73.57} \\ \hline
\multirow{4}{*}{\model}  & a group of \hl{friends} gathered together for a \hl{party} . the \hl{turkey} was delicious . it was a delicious \hl{meal} . \hl{everyone} was having a great time . after the party , we all sat down and talked about the night . my friend and \hl{[female]} were very happy to \hl{drink} . & \multirow{4}{*}{3.32} & \multirow{4}{*}{16.99} & \multirow{4}{*}{41.71} & \multirow{4}{*}{78.51} \\\hline
\end{tabular}%
}
\caption{Example stories generated by our model and BLEU-RL. Words in yellow indicate entities appearing in the image, and words in blue show repetitive patterns. Quality metrics including our proposed reward scores and BLEU-4 (R: Relevance, C: Coherence, E: Expressiveness, B: BLEU-4) are shown on the right.}
\label{fig:reco-bleu}
\end{figure*}

\subsection{Qualitative Analysis}
In Figure~\ref{fig:example}, we show two image streams and the stories generated by four models. For the second image stream on the right, \bleu repeatedly generates uninformative segments, such as {\color{blue}``we had a lot of people there''}, even though \bleu achieves high scores on automatic metrics. The same problem exists in the stories generated by HSRL in the first and second examples such as {\color{blue}``i had a great time.''}. From our observation, when the images are similar across an image stream, the three baseline methods are not able to discover the different content between subsequent images, thus generating repeated sentences with redundant information.

With regards to the relevance between the visual concepts in the image stream and the stories, \model consistently generates more specific concepts highly correlated to the appearing objects in the image stream. In Figure~\ref{fig:example}, words highlighted in yellow represent the entities that can be grounded in the images. In the second example, our \model is encouraged to generate rare entities such as ``sign'' and ``flags'' in addition to frequent entities such as ``people''.

In the first example of Figure~\ref{fig:example}, sentence pairs that are not semantically coherent are highlighted with an underline. The forth sentence generated by AREL mentions ``the president of the company'' that is quite different from the previously-described entity ``military officer'', showing that AREL forgets the content in previous images when it generates the next sentence. Similarly the second sentence generated by HSRL suddenly changes the subject of the story from ``i'' to ``he'', and mentions the ``new professor'' that is quite different from the previously-described entity ``new team''. From our observation, this type of disconnection is quite common in stories generated by the three baseline methods. The stories generated by \model are a lot more coherent in content.

Moreover, we further compare the stories generated by our proposed \model and our variant BLEU-RL. These two methods use different sentence-level reward functions during the reinforcement training. In Figure~\ref{fig:reco-bleu}, we find that \model generates more related entities such as ``meal'' and ``drink''. We also observe that the key entities generated by \model make the story more consistent in the topic, while \bleu forgets the previous context when generating the last two sentences. Our proposed \model also obtains higher scores of our proposed rewards than \bleu. 

\section{Conclusion}
In this paper, we propose ReCo-RL, a novel approach to visual storytelling, which directly optimizes story generation quality on three dimensions natural to human eye: relevance, coherence, and expressiveness. Experiments demonstrate that our model outperforms state-of-the-art methods on both the existing automatic metrics and the proposed assessment criteria. In future work, we will extend the proposed model to other text-generation tasks, such as storytelling based on some writing prompts \cite{li2019domian} and table-to-text generation \cite{table2text}.

\section{Acknowledgments}
This work was partially done while the first author was an intern at Microsoft Dynamics 365 AI Research.

\bibliographystyle{aaai.bst}
\bibliography{AAAI-HuJ.3499}

 \newpage
 \appendix

 	\section{Appendix}
     \subsection{AMT Instruction}
     We provide the template of our human evaluation on AMT.
     \subsection{Qualitative Analysis}
     We provide more examples of generated stories in the following.
    
 	\begin{figure}
 	    \centering
 		\includegraphics[width=0.8\textwidth]{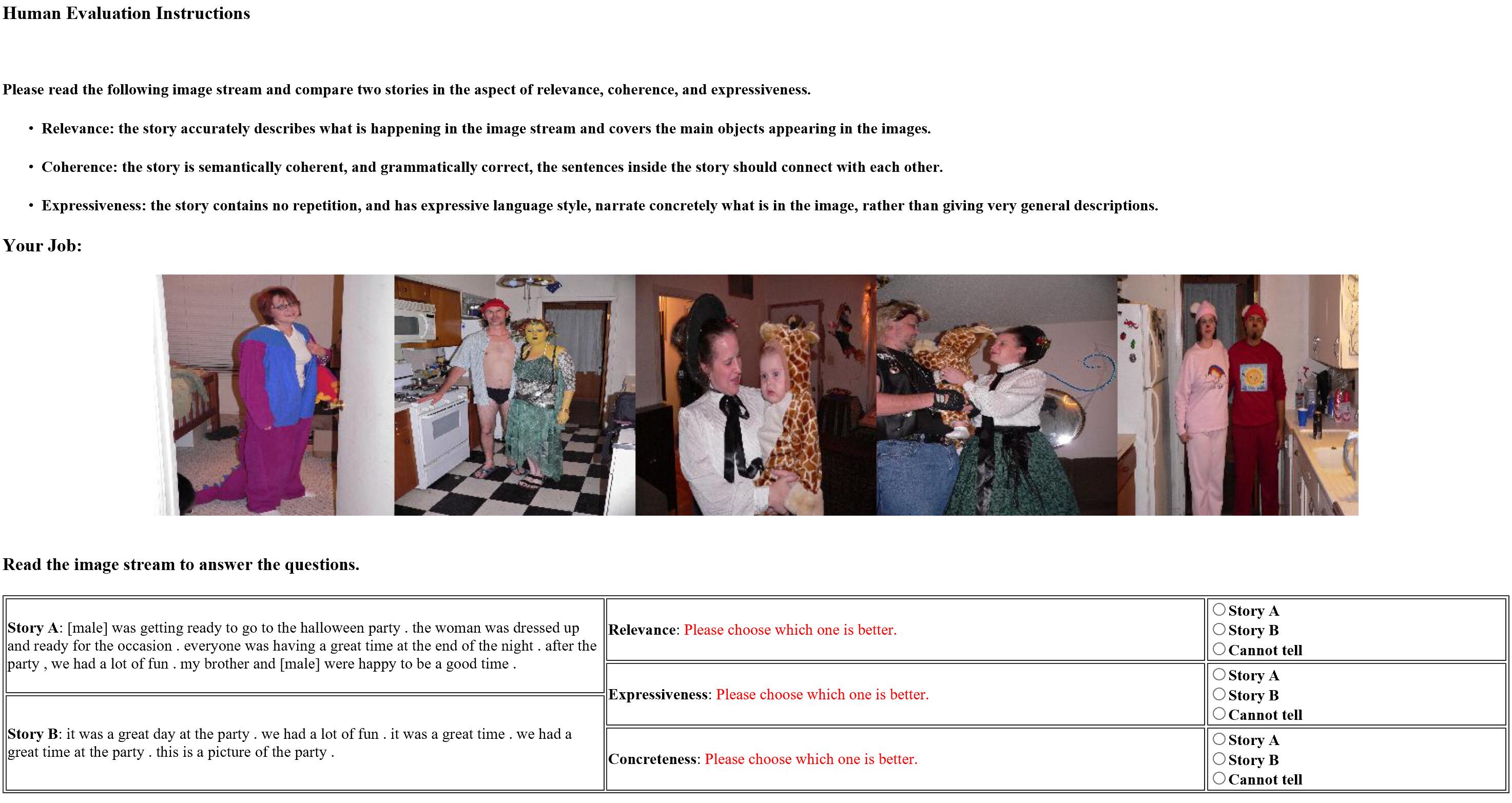}
 		\caption{AMT instruction for the user study. } \label{fig:amt}
 	\end{figure}

 \begin{figure*}
 \centering
 \raisebox{-.5\height}{\includegraphics[width=0.86\textwidth]{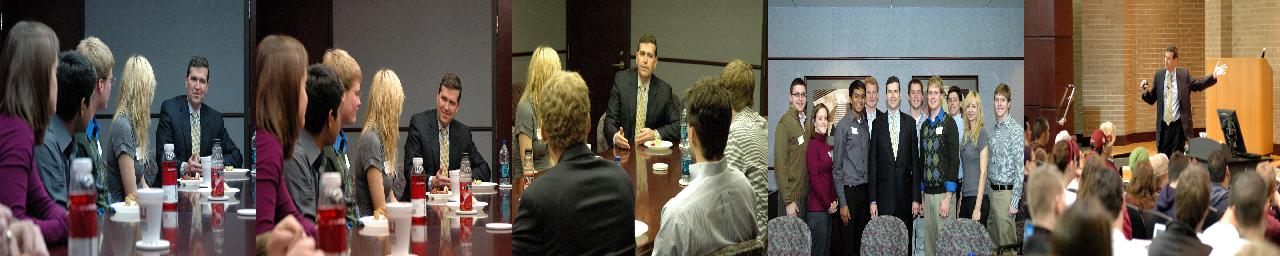}} \\
 {\small
 \begin{tabular}{|p{0.1\textwidth-2\tabcolsep}||p{0.75\textwidth-2\tabcolsep}|}  \hline
 AREL & a group of \hl{people} gathered for a \hl{meeting}. there were a lot of \hl{people} there. \underbar{the ceo} \underbar{of the \hl{meeting} was a lot of questions. the \hl{group} was very happy to be there.} the presentation was great and \hl{everyone} had a great time. \\\hline
 HSRL & the \hl{meeting} was {\color{blue}a lot of \hl{people}}. we had {\color{blue}a lot of people} there. we had a lot of \hl{questions}. {\color{blue}we had a great time. we had a great time.} \\\hline
 \bleu & a group of \hl{people} gathered for the \hl{meeting}. there were many \hl{people} there. he {\color{blue}gave a speech} about the lecture. at the end of the day, we all had a great time at the \hl{meeting}. the \hl{speaker} {\color{blue}gave a speech}. \\\hline 
 \model  & a group of \hl{people} gathered at the \hl{meeting} today. there was a lot of questions and discussed the \hl{audience}. he was very happy to see what they were there. and the \hl{speakers} were very excited to be able to take a \hl{picture} together. the \hl{speaker} was giving a \hl{speech}. \hl{he} had a great time and listened to the \hl{crowd}. \\
 \hline 
 \end{tabular} }  
 \raisebox{-.5\height}{\includegraphics[width=0.86\textwidth]{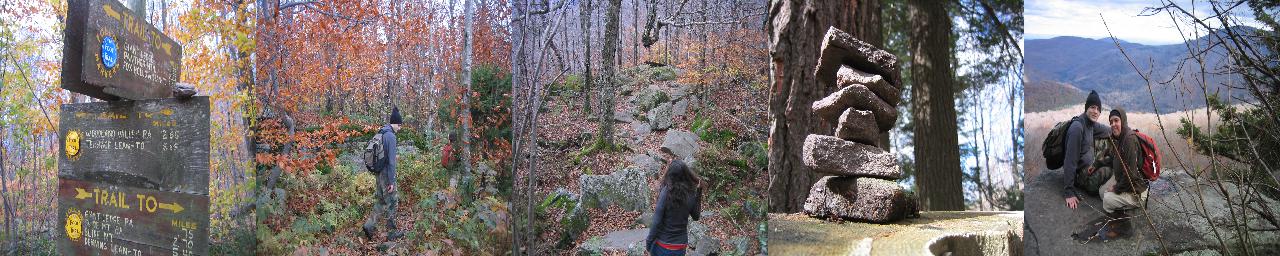}} \\
 {\small
 \begin{tabular}{|p{0.1\textwidth-2\tabcolsep}||p{0.75\textwidth-2\tabcolsep}|}  \hline
 AREL & we took a \hl{trip} to the \hl{woods}. the \hl{trees} were very beautiful. we saw a lot of cool \hl{trees}. this is a picture of a \hl{tree}. we took a lot of pictures. \\ \hline
 HSRL & for our \hl{trip} to {\color{blue}location location location}. we went to the \hl{woods} and saw some beautiful \hl{trees}. the first thing we saw was a \hl{rock climbing}. there were also a \hl{stone statue} of the \hl{rock}. {\color{blue}we had a great time} at the end of the day, {\color{blue}we had a great time.} \\\hline 
 \bleu & we went to the {\color{blue}location location location}. \underbar{we walked through the \hl{woods}. there was a} \underbar{lot of pictures.} there was a lot of interesting \hl{statues}. we had a great time. \\\hline
 \model  & we took a \hl{trip} to the location today. the \hl{forest} was a great time to take a walk through the \hl{woods}. they were so excited to see the \hl{trees} and saw a lot of fun. this is a picture of a \hl{rock formations}. it was so beautiful. \hl{my friend and [male]} took pictures of the \hl{mountain}.\\
 \hline 
 \end{tabular} }
 \raisebox{-.5\height}{\includegraphics[width=0.86\textwidth]{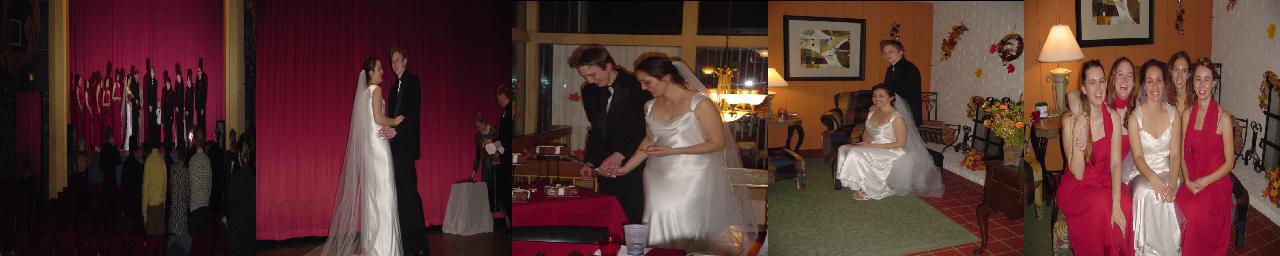}} \\
 {\small
 \begin{tabular}{|p{0.1\textwidth-2\tabcolsep}||p{0.75\textwidth-2\tabcolsep}|}  \hline
 AREL & {\color{blue}the \hl{wedding} was} held at a church. {\color{blue}the \hl{wedding} was} beautiful. {\color{blue}the \hl{bride} and \hl{groom}} cut the \hl{cake}. {\color{blue}the \hl{bride} and \hl{groom}} were very happy. the whole family was there to celebrate. \\\hline
 HSRL & the \hl{wedding} was a beautiful \hl{wedding}. the \hl{bride} and \hl{groom} cut the dance together. {\color{blue}the \hl{bride} and \hl{groom} were very happy to be married. the \hl{bride} and \hl{groom} were very happy. at the end of the night, the \hl{bride} and \hl{groom} were happy to be married.} \\\hline
 \bleu & it was a beautiful day for the \hl{wedding}. at the end of the night, {\color{blue}the \hl{bride} and \hl{groom} were very happy. the \hl{bride} and \hl{groom} were very happy.} she was so happy to be married. the \hl{bride} and groom pose for pictures. \\\hline
 \model  & it was a beautiful day at the \hl{wedding party}. the \hl{bride} and \hl{groom} were so happy to be married. [female] was happy and she was so excited to celebrate. she had a great time to take a picture of her \hl{wedding}. all of the \hl{girls} posed for pictures. \\
 \hline 
 \end{tabular} }
 \caption{Extra example stories generated by our model and the baselines. Words in yellow indicate entities appearing in the image, and words in blue show repetitive patterns. Pairs of sentences that describe different topics are annotated by an underline. } \label{fig:extra_example}
 \end{figure*}

\end{document}